\documentclass[times,twocolumn,final,authoryear]{elsarticle}
\usepackage{jasr}
\usepackage{framed,multirow}
\usepackage{amssymb}
\usepackage{latexsym}
\usepackage{url}
\usepackage{tabularx}
\usepackage{graphicx}
\usepackage{mathtools}
\usepackage{booktabs,array}
\usepackage{makecell}
\usepackage{subfig}
\usepackage{rotating}
\usepackage{lineno}
\usepackage{xcolor}

\definecolor{newcolor}{rgb}{.8,.349,.1}

\usepackage[citebordercolor=white]{hyperref}

\journal{Advances in Space Research}

\begin{document}

\verso{Hossein Bagheri}

\begin{frontmatter}

\title{A \textcolor{red}{Machine Learning-based} Framework for High Resolution Mapping of PM2.5 in Tehran, Iran, Using MAIAC AOD Data}

\author[]{Hossein \snm{Bagheri}\corref{cor1}}
\cortext[cor1]{Tel.: +98-31-3793-5299, Email: h.bagheri@cet.ui.ac.ir}

\address{Faculty of Civil Engineering and Transportation, University of Isfahan, Isfahan, Iran}


\begin{abstract}

\textcolor{blue}{This is the pre-acceptance version, to read the final version, please go to Advances in Space Research on ScienceDirect: \url{https://www.sciencedirect.com/science/article/abs/pii/S0273117722001284}}This paper investigates the possibility of high resolution mapping of PM2.5 concentration over Tehran city using high resolution satellite AOD (MAIAC) retrievals. For this purpose, a framework including three main stages, data preprocessing; regression modeling; and model deployment was proposed. The output of the framework was a machine learning model trained to predict PM2.5 from MAIAC AOD retrievals and meteorological data. The results of model testing revealed the efficiency and capability of the developed framework for high resolution mapping of PM2.5, which was not realized in former investigations performed over the city. Thus, this study, for the first time, realized daily, 1 km resolution mapping of PM2.5 in Tehran with $R^{2}$ around 0.74 and RMSE better than 9.0 $\frac{ \mu  g}{ m^{3}}$.
\end{abstract}


\end{frontmatter}


\section{Introduction} \label{intro}
In recent decades urbanization and industrialization in Tehran have exposed many people living in urban and suburban areas to dangerous air pollutants. In this regard, urban air quality monitoring is an essential matter of concern for municipal administrations and responsible public health organizations. Among different pollutant materials, particulate matters (PM) with a size of smaller than 2.5 $ \mu m $ are found to be the leading cause of cardiovascular diseases \citep{dominici2006fine}, respiratory diseases \citep{peng2009emergency}, myocardial infarction \citep{peters2001increased}, and subsequently increasing the number of morbidities \citep{lippmann2000association}, mortalities \citep{klemm2000daily}, and hospital admissions \citep{lippmann2000association}. For Tehran, Heger and Sarrat have reported that PM2.5 is the reason for 4000 deaths per year \citep{heger2018air}. Thus, accurate estimation of PM2.5 is a vital prerequisite for air quality studies and epidemiological investigations. For this purpose, air quality measuring and monitoring stations are launched that provide high temporal resolution measurements of PM2.5. However, in Tehran, these stations are located sparsely in space (see Fig. \ref{fig.tehran}), and the variation of PM2.5 concentration over space domain cannot be modeled for better exposure assessment of PM2.5. As a solution, early studies proposed using spatial interpolation such as kriging, nearest neighbor, etc., to densify PM2.5 measurements \citep{ijerph110909101,7576703,ijgi7090368}. Since different factors play roles in variation modeling of PM2.5, using merely interpolation cannot add auxiliary information for this modeling \citep{di2016assessing}. For instance, some affecting factors related to land use parameters such as road density, amount of urbanization, and others should be considered for modeling PM concentration variation \citep{BECKERMAN2013172, VIENNEAU2010688}. However, land-use terms change slightly through time and they alone are not sufficient for high resolution PM2.5 modeling \citep{HOEK20087561}.

\textcolor{red}{As a solution, satellite-based products can be applied for high resolution PM2.5 modeling \citep{SOREKHAMER2020106057}}. In this regard, aerosol optical depth (AOD) data are widely employed for PM2.5 concentration estimation \citep{https://doi.org/10.1029/2003GL018174, YOU20151156, YAO2018819}.  Wang and Christopher illustrated the dependency of AOD and PM2.5 measurements \citep{https://doi.org/10.1029/2003GL018174}. Several studies have also reported applying AOD data along with meteorological measurements for PM2.5 estimation \citep{atmos9030105, https://doi.org/10.1029/2008JD011497, https://doi.org/10.1029/2008JD011496}. Satellite sensors such as Aqua and Terra boarding on Moderate Resolution Imaging Spectroradiometer (MODIS) provide the possibility of daily AOD measurement in extensive area coverage. Two well-known AOD products provided by Aqua and Terra sensors are Deep Blue (DB) AOD and Dark Target (DT) AOD, which are named based on the algorithm of AOD retrieval. The DB algorithm is fundamentally used to retrieve AOD over bright surfaces mainly found over urban areas \citep{sayeretal, sayer2015effect}. The DT algorithm is designed to retrieve AOD over dark vegetated surfaces. Consequently, the performance of DT decreases for bright surfaces primarily found in urban areas \citep{amt-6-2989-2013}. Both products are provided daily at either 10 km or 3 km resolution.

Recently a high-resolution retrieval of AOD at 1 km resolution is provided by a new generic algorithm, the Multiangle Implementation of Atmospheric Correction (MAIAC), which has been extensively used for air quality and epidemiological studies \citep{di2016assessing, xiao2017full, liang2018maiac}. The algorithm is based on time series processing of Aqua and Terra datatakes, which separates dynamic features such as aerosols and clouds from surface properties that are relatively static during the short period \citep{amt-11-5741-2018}. The high spatial resolution output of MAIAC makes retrieved AODs a potential source for precise mapping of AOD compared to DT and DB. Mhawish et al. compared the MAIAC AOD with DB- and DT-derived AODs and demonstrated the ability of MAIAC AOD for the better revealing of air pollution sources in the south of Asia \citep{MHAWISH201912}. Regarding the correlation of AOD and PM2.5 proved in different investigations \citep{https://doi.org/10.1029/2003GL018174, https://doi.org/10.1029/2008JD011496}, MAIAC data can be used for high resolution modeling and mapping of PM2.5 variability in urban areas.
However, PM2.5 estimation from MAIAC AOD is a challenging task in the study area (Tehran), and important parameters influence the accuracy of PM2.5-AOD modeling.  

As will be illustrated in Section \ref{review}, in the literature, no study has been realized high resolution PM2.5 mapping over Tehran city in practice. Thus, this paper proposes a framework for high resolution mapping of PM2.5 using MAIAC AOD and other relevant parameters. This framework consists of 3 main stages: data preprocessing, regression modeling based on the machine learning techniques, and model deployment for daily, high resolution mapping of PM2.5. More details of the framework are presented in Section \ref{sec.frame}. 

The remainder of this paper is organized as follows. First, some related investigations are reviewed in Section \ref{review}. The study area and datasets employed in this research are introduced in Section \ref{study_area} and \ref{data}, respectively. Next, details of the devised framework including data analyzing and preprocessing, statistical PM2.5 modeling using machine learning techniques, and finally high resolution PM2.5 map generation through the model deployment are explained in Section \ref{sec.frame}. Then, the results of experiments are presented in Section \ref{result}, and in the following, the feasibility of PM2.5 mapping over the study area is discussed. Section \ref{sec.conclusion} presents the conclusions of this study.

\section{Related Work} \label{review}

\textcolor{red}{Three main types of models have been developed for PM2.5 concentration estimation from satellite AOD measurements: Chemical simulation models \citep{https://doi.org/10.1029/2004JD005025, van2010global}, statistical models \citep{ma2016satellite, song2014satellite}, and semi-empirical models \citep{LIN2015117}. Among them, statistical models are more popular to implement for PM2.5 modeling, and machine learning techniques have been widely used for this type of modeling.
In the literature, simple linear regression models (univariate or multivariate) have been accomplished for PM2.5 concentration estimation. In addition to linear regression models, advanced machine learning algorithms have also been applied for PM2.5 concentration estimation \citep{atmos9030105, https://doi.org/10.1029/2008JD011496, https://doi.org/10.1029/2008JD011497, https://doi.org/10.1002/2017GL075710, AHMAD2019117050, chen2020estimating, SUN2021144502}. For example, Gupta and Christopher designed a multi-layer perceptron (MLP) to explore the relationship between AOD and PM2.5 using meteorological data \citep{https://doi.org/10.1029/2008JD011496}. Li et al. developed a geointelligent network using a deep belief Boltzmann structure for estimating PM2.5 \citep{https://doi.org/10.1002/2017GL075710}. Other machine learning algorithms such as support vector regressor (SVR) \citep{vapnik2013nature}, random forest \citep{james2013introduction}, gradient boosting \citep{friedman2002stochastic}, etc., have been used for estimating PM2.5 concentration from meteorological data and AOD as input features. \citeauthor{Weizhen_2014} developed a successive over relaxation SVR model using Gaussian kernel function for predicting PM2.5 and PM10 by satellite AOD and meteorological parameters in Beijing. The decision tree ensemble approaches have been broadly used for modeling PM concentration from AOD retrievals \citep{SUN2021144502, LU2021100734, hu2017estimating, CHEN2021110735, YANG2020111061, JIANG2021105146}. \citeauthor{LU2021100734} trained random forest to predict the PM2.5 level over several urban areas in China using high resolution AOD and meteorological data \citep{LU2021100734}. In another study, land use data and column water vapor in addition to AOD and meteorological parameters were involved for high resolution mapping of PM2.5 \citep{doi:10.1021/acs.est.0c01769}. In this study, the PM2.5 level was predicted by a linear mixed effect model and random forest \citep{doi:10.1021/acs.est.0c01769}.
	XGBoost as a gradient boosting approach is another approach utilized for PM2.5 modeling by AOD data. For example, \citeauthor{rs12203368} designed a spatially local extreme gradient boosting (SL-XGB) model for PM2.5 prediction from SARA AOD at urban scales \citep{rs12203368}.}

\textcolor{red}{In addition to classical methods, neural networks as another category of popular machine learning techniques have been utilized for PM2.5 estimation from AOD data \citep{atmos9030105, https://doi.org/10.1029/2008JD011497}. 
In recent years, deep neural networks have proved their performances in different tasks of classification and regression. For PM2.5 estimation from satellite data, deep learning techniques have also been applied and compared with classical machine learning approaches.  The efficiency of deep neural network structures has been illustrated in several investigations \citep{WANG2019128, rs12020264}. \citeauthor{rs12020264} used autoencoder-based residual networks for estimating PM2.5 and PM10 from AODs \citep{rs12020264}. \citeauthor{CHEN2021144724} used a self-adaptive deep neural network for finding the PM2.5-AOD relationship \citep{CHEN2021144724}.}

\textcolor{red}{In the literature, the successful PM2.5 modeling from AOD data in Tehran has been mainly realized based on the 3 or 10 km (DB or DT) MODIS products. Earlier studies mainly focused on PM2.5 concentration estimation using satellite-based AOD measurements at lower resolution (\texttildelow 10 km) in Tehran. In an earlier study, it was tried to estimate PM2.5 using 10 km DT AODs in a short period of observations. The correlation of predicted PM2.5 and observed PM2.5 was around 0.55 \citep{sotoudeheian2014estimating}. In another investigation, Ghotbi et al. could estimate PM2.5 over Tehran with higher accuracy ($ R^{2} $ = 0.73) using the 3 km DT AOD and meteorological data derived from climate stations \citep{GHOTBI2016333}. However, they used few samples (332 data points) collected from few stations for a very short period from March to November 2009. Another study attempted to estimate PM2.5 from 10 km MODIS AOD (combined DB and DT) product over Tehran. The results demonstrated that using machine learning techniques gave accuracy up to 80\% \citep{atmos10070373}. PM2.5 estimation from high resolution satellite imagery such as Landsat satellite imagery has been investigated in several studies \citep{jafarian2020evaluation, IMANI2021111888}. However, these images have a lower temporal resolution (e.g., image acquisition per every six days ), which is not suitable for daily representation of the PM2.5 map. Only a study performed over Tehran using MAIAC AOD data has reported a correlation of less than 0.5, on average \citep{NABAVI2019889}, which is not perfect for high resolution PM2.5 mapping based on AOD.}
	
\textcolor{red}{Review of previous investigations studied on Tehran urban area demonstrates that there is no practical implementation for daily, high resolution mapping of PM2.5 concentration  over the study area. Consequently, the main focus of this paper is to develop a framework based on machine learning to reach the goal of high resolution PM2.5 mapping over the study area.}

\section{Study Area and Materials}
\subsection{Study Area}\label{study_area}

\begin{figure*}[t!]
	\begin{center}
		\includegraphics[width=0.7\textwidth]{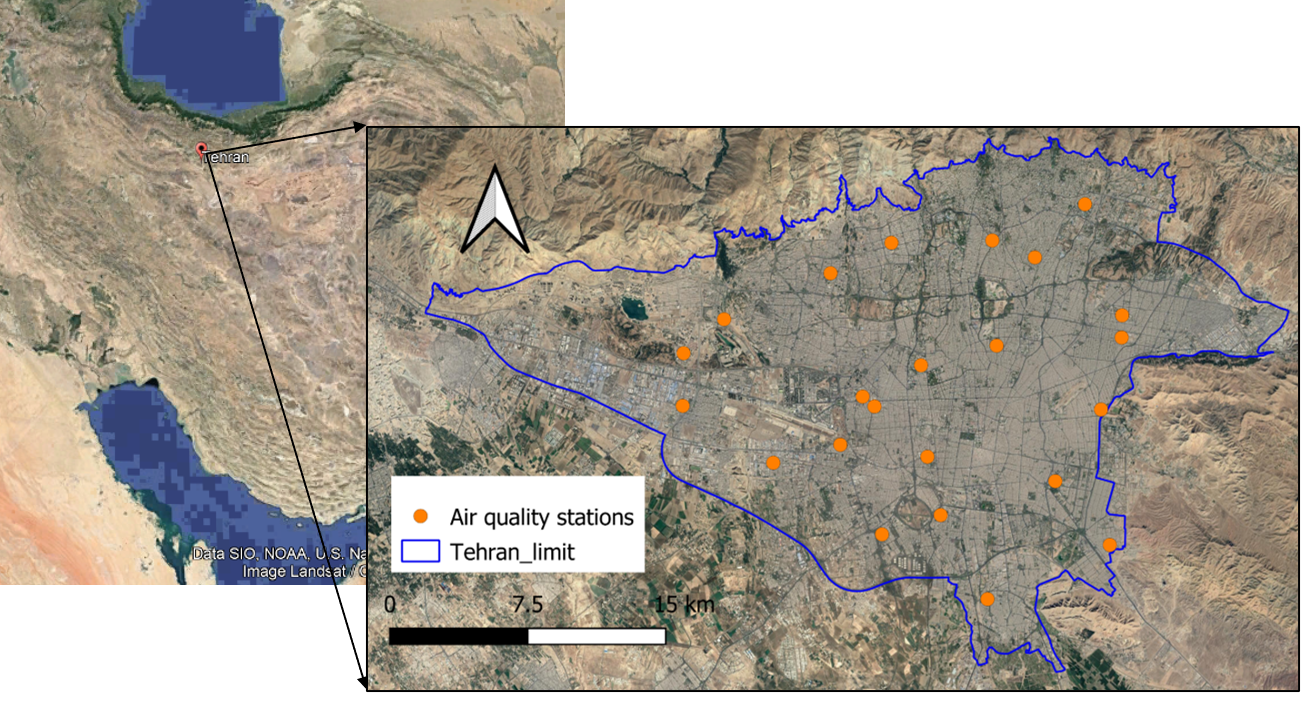}
		\caption{Visualization of the study area, Tehran, the capital of Iran. Locations of air quality monitoring stations are marked by orange circles.}
		\label{fig.tehran}
	\end{center}
\end{figure*}

The study area is Tehran city, the capital of Iran (shown in Fig. \ref{fig.tehran}), with a population of 13.3 million residents and 10 million commuters. It spreads from latitude  35$ ^{\circ} $ 35$'$ N to  35$ ^{\circ}  $48$'$ N and longitude  51$ ^{\circ}  $17$'$ E to  51$ ^{\circ} $ 33$'$ E. The highest point of the city has an elevation of 1800 m, while the lowest height is more than 900 m above the mean sea level. One of the primary sources of pollution is mobile sources such as vehicles and a relatively old fleet, which produce around 85\% of the total pollutants and 70\% of PM \citep{ARHAMI201770}. Also, human activities such as changing the land use and land cover of the urban and suburban areas increased the intensity of air pollution. Due to specific mountainous topography \textemdash surrounding the city by mountains from the north to the southeast\textemdash winds carry the air pollution from the industries in the west of the city to the middle and the east \citep{ATASH2007399}. 

\subsection{Materials}\label{data}
For this study, different datasets, AODs collected by satellite; meteorological data provided by the global weather model; PM2.5 measured at ground air monitoring stations, are utilized. The datasets were collected for seven years, from Jan. 2013 to Jan. 2020. In the following, more details of each dataset are described.

\subsubsection{PM2.5 Monitoring Data}\label{pm2.5}
The mean daily PM2.5 level is collected by Tehran's Air Quality Control Company (AQCC). Fig. \ref{fig.tehran} shows the locations of the air quality monitoring stations. As shown in Fig. \ref{fig.tehran}, the air quality of the city is monitored by 23 stations scattered across the city. PM levels are measured hourly by a Tapered Element Oscillating Microbalance (TEOM) instrument \citep{sotoudeheian2014estimating}. Despite the good spread of monitoring stations, they are not sufficient for high resolution PM2.5 mapping of the city. Interpolation techniques ignore the variability of weather situations and human-made factors such as local emissions. The case becomes worse when the continuous measurement of PM2.5 by existing stations is not possible since, over time, some stations become out of order which means reducing the number of air monitoring stations.  For example, among those 23 stations, measurements of some stations are not available in some periods due to technical issues.

\subsubsection{MAIAC AOD}\label{AOD}
AOD can be employed to model the variability of PM2.5 levels in locations between monitoring stations. Spaceborne sensors can provide daily AOD measurements. AOD identifies the columnar aerosol level in the atmosphere by measuring the light extinction induced by aerosols. \textcolor{red}{Two widely used satellite AOD products are those retrieved by DB and DT algorithms from MODIS Aqua and Terra sensors. While the DT algorithm is mainly applicable for dark vegetated areas, which restricts its usage in urban areas \citep{amt-6-2989-2013}, the MODIS DB algorithm was originally developed to retrieve AOD over bright surfaces using 470 and/or 412/650 nm, depending on the surface \citep{sayeretal}. The second generation C6 version of the DB product has been further updated by  \citeauthor{https://doi.org/10.1002/jgrd.50712} considering an improved assessment of NDVI-dependent surface reflectance, improved cloud screening and identification of dust. This helped to extend the applicability of the DB algorithm from the arid/desert region to the entire land surface except for snow/ice-covered areas \citep{https://doi.org/10.1002/jgrd.50712}.}
Nevertheless, both algorithms lead to AOD data at 10 km or 3 km resolution, which limits a high-resolution PM monitoring, particularly over urban areas.

A recent AOD product is the output of the MAIAC algorithm that uses time series of measurements acquired by Aqua and Terra sensors boarding on the MODIS satellite platform. The algorithm gives an AOD product at a resolution of 1 km which can be applied for high-resolution mapping of PM2.5, especially over urban areas \citep{amt-11-5741-2018}. \textcolor{red}{In this study, the MCD19A2 Version 6 of MAIAC data product is employed for PM2.5 estimation.}

\subsubsection{Meteorological Data}\label{met}
In addition to AOD, several investigations demonstrated the significance of meteorological data for PM2.5 concentration estimations \citep{https://doi.org/10.1029/2008JD011496, https://doi.org/10.1029/2008JD011497,atmos9030105}. The meteorological data can be collected by either weather stations or be provided by weather models. 

One of the famous global weather models that can provide uniformly distributed meteorological data around the whole world is the model developed by European Centre for Medium-Range Weather Forecasts (ECMWF). The meteorological data can be gathered from the fifth-generation ECMWF reanalysis for the global climate and weather, namely, ERA5 \citep{ER5}. ERA5 estimates the atmospheric, ocean-wave, and land-surface quantities hourly \citep{https://doi.org/10.1002/qj.3803}. It combines model data (a previous forecast) and newly available observations to update the estimate of the atmosphere. Another version of ERA5 is ERA5-Land hourly data that provides land variables at enhanced resolution comparing to ERA5 \citep{ER5-land}. All required meteorological data used in PM2.5 concentration estimation can be derived from ERA5 and ERA5-land hourly data. 

Tab. \ref{feature} expresses the characteristics and source of meteorological data used in this study for PM2.5 estimation. As expressed in Tab. \ref{feature},  both versions of ERA5 models can provide the meteorological data in a grid format with a higher spatial resolution compared to synoptic stations’ observations, which can potentially be employed for high resolution mapping of PM2.5. 
\begin{table*}[bt!]
	\centering \footnotesize
	\caption{Input features used for PM2.5 modeling in urban areas. Note that the column Notation represents the notations of features used in this paper. }
	\label{feature}
	\begin{tabular}{llcc}
		 Notation &Description & Data Source & Resolution  \\\hline
        \toprule
        
		 AODm & Mean aerosol optical depth	& MAIAC  & 1 km \\ 
		 nAODm & Normalized	mean AOD & \thead{MAIAC \\ ECMWF} & 1 km \\
		 Prob\_bestm & Probability of mean AOD to have best quality & MAIAC  & 1 km \\
		 Prob\_medm & Probability of mean AOD to have medium quality & MAIAC  & 1 km \\
		 lat & Latitudinal position of the air quality monitoring station	& \thead{MAIAC \\ AQCC} & 1 km \\
		 long & Longitudinal position of the air quality monitoring station	& \thead{MAIAC \\ AQCC} & 1 km \\
		 d2m & 2m dewpoint temperature	& ECMWF &  \texttildelow 10 km \\
		 t2m & 2m temperature	& ECMWF &  \texttildelow 10 km \\
		 blh or PBLH& Planetary boundary layer height	& ECMWF & \texttildelow 10 km \\
		 sp & Surface pressure	& ECMWF &  \texttildelow 10 km \\ 
		 lai\_hv & Leaf area index, high vegetation	& ECMWF &   \texttildelow 10 km \\
		 lai\_lv & Leaf area index, low vegetation	& ECMWF &   \texttildelow 10 km \\ 
		 ws10 & 10m wind speed	& ECMWF &\texttildelow 10 km \\ 
		 wd10 & 10m wind direction	& ECMWF &  \texttildelow 10 km \\
		 cdir & Clear sky direct solar radiation at surface	& ECMWF & \texttildelow 10 km \\
		 uvb & Downward UV radiation at the surface	& ECMWF &  \texttildelow 10 km \\ 
		 RH & Relative humidity	& \textcolor{red}{ECMWF} & \texttildelow 10 km\\
		 month & Month 	& --- & --- \\
		 DOY & Day of year	& --- & --- \\	
		\hline		
	\end{tabular}
	
\end{table*}

\section{A Framework for High Resolution PM2.5 Mapping Using MAIAC AOD}\label{sec.frame}
\begin{figure*}[t!]
	\begin{center}
		\includegraphics[width=1\textwidth]{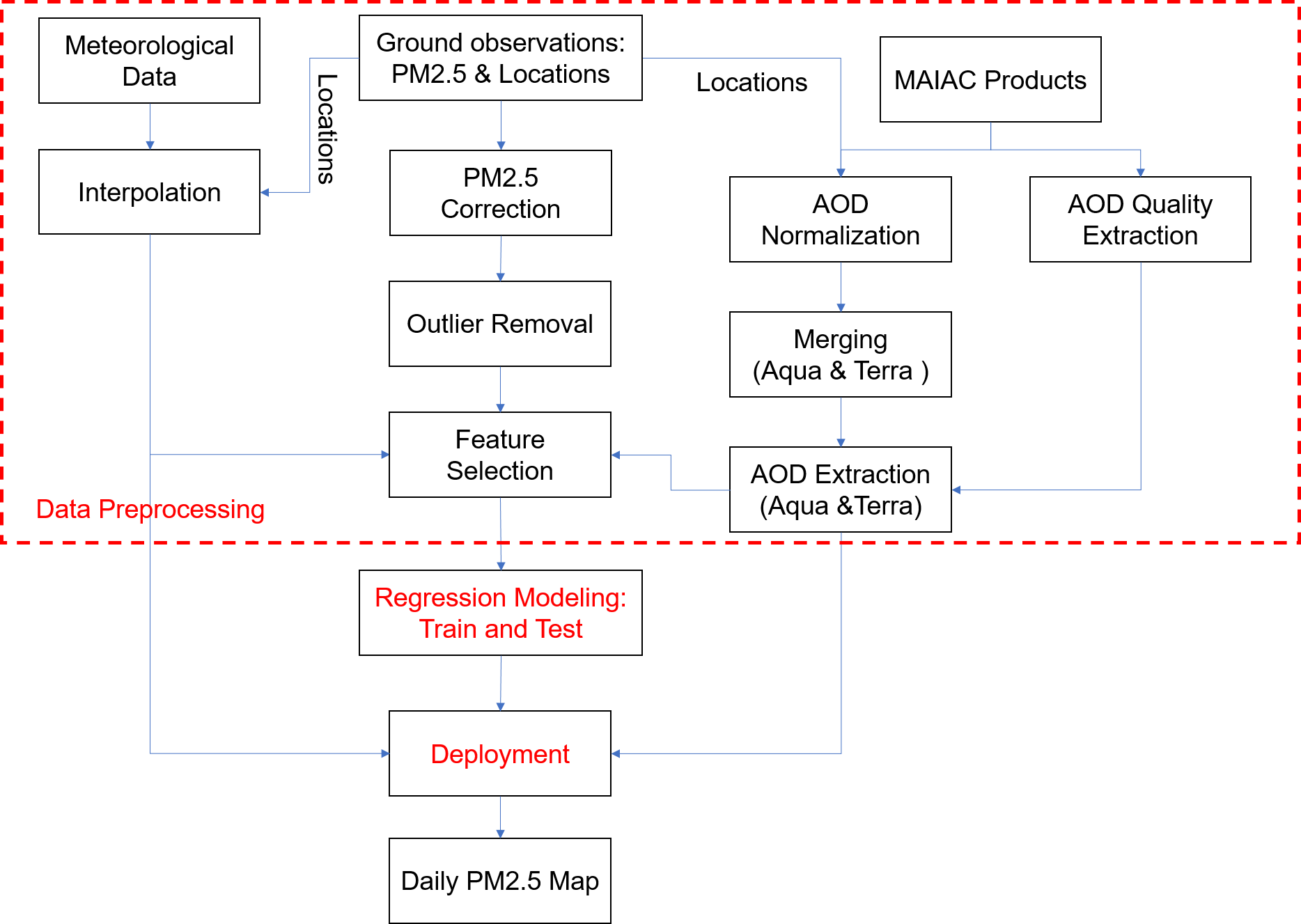}
		\caption{The framework devised for high resolution estimation of PM2.5 over Tehran}
		\label{fig.frame}
	\end{center}
\end{figure*}

Fig. \ref{fig.frame} displays the devised framework suited to Tehran city for high resolution estimation and mapping of PM2.5 using AOD, meteorological data, and other features. The framework consists of three main stages, data preprocessing; regression modeling; and deployment. First, data become prepared in the preprocessing phase. In other words, the objective of this stage is to prepare features for importing into the next module i.e., regression modeling. In the regression modeling module, a machine learning technique is developed to explore the relationship between input features (AOD, meteorological data, etc.) and corresponding PM2.5 collected at ground stations. The achieved model from the regression is employed in the deployment stage to finally produce daily high resolution PM2.5 maps over the study area. More details of each step and embedding modules are described in the following sections. 

\subsection {PM2.5 Data Preprocessing}
\subsubsection{PM2.5 Correction}\label{correctionPM}
Earlier studies illustrated the relationship between AOD retrieved by different algorithms from MODIS observations and PM2.5 measured by air quality monitoring stations mainly based on the univariate linear regressor. For example,  Wang and Christopher showed a correlation of  0.76 and 0.67 between PM2.5 values and AOD products derived from Aqua and Terra, respectively \citep{https://doi.org/10.1029/2003GL018174}.  Despite the good correlation between AOD and PM2.5 in the mentioned study, several studies have implied that this relationship could be significantly affected by the vertical distribution of aerosols and the ambient relative humidity \citep{tsai2011analysis, zhang2016semi, engel2006integrating, wang2010satellite}. 
In Tehran, PM2.5 and PM10 collected at monitoring stations are measured by TEOM after heating the ambient air to 50$ ^{\circ} $C \citep{sotoudeheian2014estimating, GHOTBI2016333}, and consequently, the mass of dry PM reported as measured PM is less than raw PM. This correction can be performed as below \citep{tsai2011analysis}:

\begin{equation}
PM_{c}= PM (1-\dfrac{RH}{100})^{-1}, \label{pm}
\end{equation}
where $ PM_{c} $ is the corrected value of measured $ PM $ at the monitoring station, and RH is the relative humidity.


%
%

\subsubsection{Outlier Removal}\label{sec.outlier}
PM2.5 values used in the study are daily averages of 24-hours PM2.5 measurements at air quality monitoring stations. A daily PM measurement is the average of at least 80\% of hourly valid data in a day recorded at each station and below this percent is reported as missing. While an averaging decreases the effect of possibly existing noise or outliers in hourly measurements, some hourly measurements may dramatically deviate from the actual values. In this case, even averaging cannot degrade deviations. Thus, these types of measurements are considered outliers. In this paper, two simple strategies are carried out for outlier detection and removal. First, the interquartile range (IQR) is assumed to separate inlier measurements from outliers \citep{yang2019outlier}. In this way, the inliers are obtained by the condition below: 

\begin{equation}
	Q_{1}-IQR < PM2.5<Q_{3}+IQR,
\end{equation}
where $ Q_{1} $ and $ Q_{3} $ are the first and third quartiles of input PM2.5 and $ IQR $ is the interquartile range of PM2.5 values. 

The second strategy is based on the standard deviation of input PM2.5, which is called $  3\sigma $ in this paper \citep{posio2008outlier, bagheri2018fusion}. The inlier PM2.5 measurements are those that 

\begin{equation}
	\mu-3\sigma<PM2.5<\mu+3\sigma,
\end{equation}
where $ \mu $ and $ \sigma $ are the mean and standard deviation of PM2.5 measurements.

\subsection{MAIAC AOD Data Preparation}

\subsubsection{AOD Normalization}
Another required modification is to normalize MAIAC AOD data. Since AOD is a columnar parameter while the PM values are measured at the surface nearby the station, a conversion from the columnar to the surface AOD measurement is necessary \citep{wang2010satellite}. In this regard, original AOD values should be normalized before any further processing \citep{tsai2011analysis}. This can be achieved by the height of the mixing layer at each monitoring station. Thus, the normalized AOD is calculated as \citep{tsai2011analysis}:

\begin{equation}
	nAOD = \dfrac{AOD}{L_{mix}}, \label{naod1}
\end{equation} 
where $ nAOD $ and $ AOD $ are the normalized AOD and the original AOD values retrieved by the MAIAC algorithm, respectively, and $ L_{mix} $ denotes the mixing layer height. In this study, it is assumed that aerosols are homogeneously mixed and the height of the haze layer is ignored in normalization. In addition, a previous investigation over Tehran city disclosed that aerosol layer height (ALH) (derived from CALIPSO profiles over the study area) and the planetary boundary layer height (PBLH) have the same altitude above the aerosol-laden layers \citep{NABAVI2019889}. As a result, $ L_{mix} $ can be replaced with PBLH. Therefore, eq. \ref{naod1} can be updated as below:

\begin{equation}
	nAOD = \dfrac{AOD}{PBLH}, \label{naod2}
\end{equation} 
where $ PBLH $ is the planetary boundary layer height obtained from ECMWF model.

\subsubsection{AOD Extraction from MAIAC Products}\label{aod_extraction}

AOD data provided by MODIS MAIAC is initially in a raster format, while for the statistical modeling of PM2.5, AOD is extracted at each air monitoring station. To obtain the coincident MODIS pixels with the PM2.5 measurement at the monitoring station, different window sizes, 3$ \times $3, 5$ \times $5, 7$ \times $7, 11$ \times $11, 15$ \times $15 are applied to evaluate the relationship between AOD and PM2.5 values. The final AOD is the average of AOD values (AODm) inside the considered window. 

For the experiment, several criteria are considered to ensure preserving the quality of AODs after averaging. The criteria are associated with the quality of AOD values extracted from MAIAC products at each window. In more detail, after averaging AODs inside the window, the standard deviation of the AODs is calculated. If this value is more than 0.5, the achieved mean AOD will be considered an invalid value, which means AOD values in neighborhoods fluctuate severely. Since in a window, some AODs are not available (filled by NaN), another criterion is that the number of pixels with valid AOD values should be more than three, which makes the averaging of AODs more meaningful. The aforementioned criteria are considered for any  selected window sizes.

\subsubsection{AOD Quality Extraction}\label{quality_extraction}
In addition to the criteria mentioned in the previous section, other conditions can be considered using the information provided in the “Quality Assessment” (QA) file delivered along with MAIAC AOD products \citep{lyapustin2018modis}. 
According to the manual of MAIAC AOD product, and based on previous investigations \citep{just2015using, kloog2014new}, a recommendation is to merely apply those AODs for urban air quality applications that satisfying the condition below: 
 
\textbf{Condition 1}: (Adjacency Mask == Normal condition/Clear) \textbf{and} (Cloud Mask == Clear or Possibly Cloudy),
where Adjacency Mask gives information of recognized neighboring clouds or snow (in the 2-pixel vicinity).

The condition mentioned above can become stricter by filtering those AODs that are flagged as “Best quality” in the QA file \citep{lyapustin2018modis}. Consequently, the second condition is considered as:

\textbf{Condition 2}: (Adjacency Mask == Normal condition/Clear) \textbf{and} (Cloud Mask == Clear) \textbf{and} (QA for AOD == Best quality)

Regarding the fact that each window may include AODs with different qualities, the final AOD can be calculated by averaging only AODs satisfying the condition 1 or 2. However, this strategy can lead to missing valuable AODs that do not meet the conditions. 
Instead of filtering AODs based on the conditions as mentioned above, which may lead to missing the AOD information at an air monitoring station, two probability maps are generated based on those defined conditions. In this manner, AODs inside a window are averaged, and corresponding to achieved AOD, a probability representing the number of pixels (AODs) satisfying the relevant condition respective to the total number of pixels inside the window is calculated. In other words, the assigned probability illustrates the number of pixels with the highest quality (satisfying condition 2) or with medium quality (consistent with condition 1) involving in the calculation of mean AOD. These probability values can be used for controlling the quality of the final achieved AOD at each monitoring station. In this paper, two generated weight maps are notated as "Prob\_medm" and Prob\_bestm" regarding conditions 1 and 2, respectively.  

\subsubsection{Merging AODs of Aqua and Terra}\label{merg-AT}

Hu et al. \citep{hu2014estimating} and Lee et al. \citep{lee2011novel} have shown averaging AOD values retrieved by Aqua and Terra overpassing at different local times (around 10:30 and 13:30 local times) can be applied as a daily AOD measurement. 
The correlation between Aqua and Terra AODs also allows filling missing AOD values of a sensor using AODs retrieved by another sensor.  In locations where either Terra or Aqua AOD ($ AOD_{T} $ or $ AOD_{A} $) is missing, the missing value can be estimated using the computed regression equations. Then, the final AOD can be calculated as below: 
\begin{equation}
	AOD = \dfrac{AOD_{A} + AOD_{T}}{2}, \label{eq.AT}
\end{equation} 
in which missing $ AOD_{A} $ or $ AOD_{T} $ can be estimated using coefficients achieved by linear regression.

\subsection{Meteorological Data Preparation}

For employing the meteorological data, it is needed to estimate them at locations of air quality monitoring stations. For this purpose, the meteorological values are interpolated at target locations using an interpolation technique such as kriging \citep{ijgi7090368, olea2012geostatistics,bagheri2014}. Two popular types of kriging that can be used for meteorological data interpolation are ordinary and universal kriging. Besides the kind of kriging, another critical parameter that should be correctly set is the semivariogram type.  Versatile semivariograms have been designed such as linear, spherical, Gaussian, and power that are typically selected based on the study data \citep{bagheri2014, aretouyap2016lessening}. One strategy for setting the aforementioned hyperparameters is a grid search with cross-validation in which a subset of data is used to estimate those parameters. In the grid search strategy, the interpolation is done on the subset of data as training data by applying different parameters and the performance of interpolation is evaluated based on another subset of data as a validation dataset. Then, the hyperparameters are determined according to a set of parameters that gives the highest performance. 

\subsection{Feature Selection}
As illustrated in Tab. \ref{feature}, several features, including those extracted from MAIAC products, meteorological data derived from ECMWF models, etc., are input into a predictive model for predicting PM2.5 concentration. However, before establishing a regression model, selecting the most important feature will be beneficial. This procedure gives an insight into the relationship between input variables and the output target (PM2.5), which can lead to reducing non-significant features, and in some cases improving the model accuracy. For this aim, different machine learning techniques such as random forest and gradient boosting can be applied.

In this paper, as will be illustrated in Section \ref{sec.reg_result}, gradient boosting will be used as a machine learning technique for AOD-PM2.5 modeling. Additionally, It provides an ability for feature importance determination. The importance is estimated for an individual decision tree by the amount that each feature split point makes better performance, weighted the number of data, the node has observed. The average of all feature importance across all of the decision trees, called gain, identifies the final importance \citep{xu2014gradient}. 

\subsection{Regression Modeling}\label{sec.model}

Besides data, another aspect of the designed framework is the type of model performed for PM2.5 concentration estimation using AOD and meteorological data.
This paper also compares different machine learning algorithms to estimate PM2.5 from MAIAC AOD and ECMWF meteorological data. For this aim, different algorithms from the basic to advanced algorithms are carried out, and their performances are compared. In this regard, four types of machine learning algorithms, linear methods (univariate, multivariate, ridge, lasso); kernel methods (SVR);  decision tree ensemble approaches (random forest, extra trees, XGBoost); and deep neural networks (deep autoencoder+SVR, deep belief network) are implemented for exploring the relationship between input features and PM2.5 values.

\subsection{Model Deployment}
The achieved model from the regression modeling phase can estimate PM2.5 in an arbitrary location using the input features, which will lead to high resolution mapping of PM2.5. The main challenge for estimating PM2.5 is missing AOD values because of cloud contamination or failure of the applied algorithm in retrieving AOD. However, the available AODs, although few numbers, can be employed and estimate PM2.5 in addition to those values measured by an air quality monitoring station. In other words, the estimated PM2.5 values using AODs and other features ultimately generate PM2.5 in locations where have not been sensed by an air quality monitoring station beforehand. The estimated PM2.5 values can be utilized as extra measurements in addition to ground station measurements for producing a high resolution map of PM2.5. 
The produced PM2.5 data can be supposed as new PM2.5 measuring stations (quasi-stations) and thus be combined with actual stations to indicate PM2.5 variations for higher resolution mapping better. Finally, a high resolution daily map of PM2.5 is produced by an interpolation technique using all estimated and observed PM2.5 values. Additionally, monthly and yearly high resolution maps of PM2.5 can be generated using the produced daily maps by median averaging. It should be noted that all prepossessing procedures, mentioned earlier, are performed on AOD and meteorological data to make them ready for PM2.5 estimation using the developed regression model.  

\section{Results and Discussion}\label{result}
In this section, results of several experiments performed to investigate the efficiency of different modules of the proposed framework for PM2.5 concentration estimation are presented and discussed. Different metrics were employed for evaluating achieved results. Some standard metrics used in this study were root mean square error (RMSE), mean absolute error (MAE), and Pearson correlation coefficient ($ R^{2} $). 

\subsection{Data Preprocessing and Preparation}
\subsubsection{Impact of AOD Normalization}\label{res.AOD_modif}

The results of PM2.5 estimation using univariate regression model by original AODs and also normalized versions are presented in Tab. \ref{Tab.aodnormal}. The results illustrate that the modification of AOD using PBLH can significantly improve the estimations. 

\begin{table}[bt!]
	\centering \footnotesize
	\caption{a) Univariate column illustrates the results of using the original AOD and also the normalized version of AOD by PBLH. b) Multivariate column shows the effect of adding meteorological data in addition to AOD values for predicting PM2.5.}
	\label{Tab.aodnormal}
	\begin{tabular}{l |ccc|ccc}
		
		&  \multicolumn{3}{c|}{Univariate}& \multicolumn{3}{c}{Multivariate} \\
		AOD type&\thead{RMSE \\  $\frac{ \mu  g}{ m^{3}}$}& \thead{MAE \\  $\frac{ \mu  g}{ m^{3}}$} & \textbf{$ R^{2} $}&\thead{RMSE \\  $\frac{ \mu  g}{ m^{3}}$}& \thead{MAE \\  $\frac{ \mu  g}{ m^{3}}$} & $ R^{2} $ \\\hline
        \toprule
		AOD 	          & 18.53 &	15.26 &	0.01 & 11.76 & 9.35 & 0.56\\
		nAOD (normalized) & 13.78 &	10.91 &	0.40 & 11.00 & 8.64 & 0.61 \\
	\end{tabular} 	
\end{table}

\subsubsection{Results of Merging AODs of Aqua and Terra}\label{res.merg-AT}
\begin{figure*}[bt!]
	\centering
	\subfloat[]{%
		\includegraphics[width=1\columnwidth]{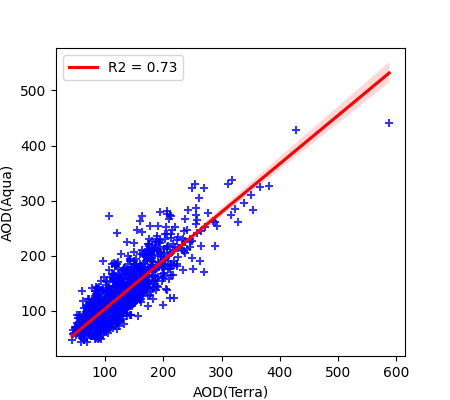}}
	\label{TA2013}
	\subfloat[]{%
		\includegraphics[width=1\columnwidth]{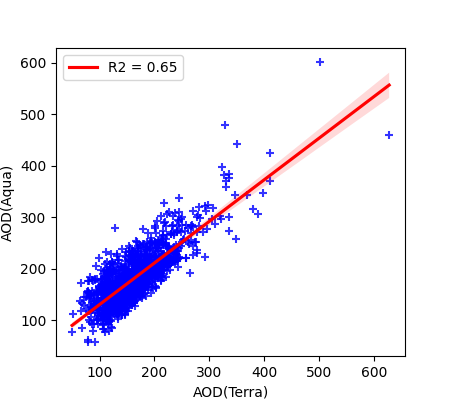}}
	\label{TAt}
	
	\caption{Correlation between Aqua and Terra AODs; a) Cold season, b) Warm season.}
	\label{fig.TA} 
\end{figure*}

For the study area in this paper, it was illustrated that a combination of AOD measurements from Aqua and Terra could be used to achieve mean daily AOD. Fig. \ref{fig.TA} displays the linear correlation between AODs retrieved from Aqua and Terra sensors for the study years from Jan. 2013 to Jan. 2020. Tab. \ref{Tab.AT} also represents the correlation coefficient as well as the linear regression equation between the AOD measurements of two sensors.  As presented in Tab. \ref{Tab.AT}, the correlation between measurements of two sensors is 0.72 when considering all measurements from Jan. 2013 to Jan. 2020. In addition, the influence of seasonality on correlation estimation between Aqua and Terra AOD has been evaluated. In this manner, the AOD values were divided into two categories; warm season (Apr. – Sep.), and cold season (Oct. – Mar.) based on the climate of the study area. Then the regression was performed for each seasonal category. The results revealed that in the cold season, the correlation was slightly higher than the case when all data were involved in regression. However, the correlation decreased for the warm season (Fig. \ref{fig.TA}). The highest correlation coefficient between AODs of Aqua and Terra is for cold season (around 0.73), when the congestion of pollution as well as missing AODs due to cloud coverage increases and accurate regression of AODs is more desirable.

\begin{table*}[bt!]
	\centering \footnotesize
	\caption{Regression equations and coefficients for retrieving missing AOD values from one sensor to another, Terra to Aqua: when Terra AOD is available, and Aqua AOD is missing; and Aqua to Terra applies vice versa. 
    }
	\label{Tab.AT}
	\begin{tabular}  {ll ccc}
	AOD distinguishing	&Time &Terra to Aqua & Aqua to Terra & $ R^{2} $   \\\hline
		\toprule
		
	\multirow{2}{*}{Seasonal}	& Cold (Oct. - Mar.)& $ AOD_{A}= 0.83AOD_{T}+21.06 	$ & $ AOD_{T}= 0.88AOD_{A}+15.47 $ & 0.73  \\
		& Warm (Apr. - Sep.)& $ AOD_{A}= 0.81AOD_{T}+15.81 $ & $ AOD_{T}= 0.81AOD_{A}+49.94 $ & 0.65  \\
	No separation	&Total (2013-2019) & $ AOD_{A}= 0.79AOD_{T}+23.39 $	& $ AOD_{T}= 0.91AOD_{A}+21.89 $ & 0.72 \\
		
		\hline		
	\end{tabular}
\end{table*}
	
\subsubsection{Window Size for AOD Extraction}\label{res.ws}
	
\begin{figure}[bt!]
	\begin{center}
		\includegraphics[width=1\columnwidth]{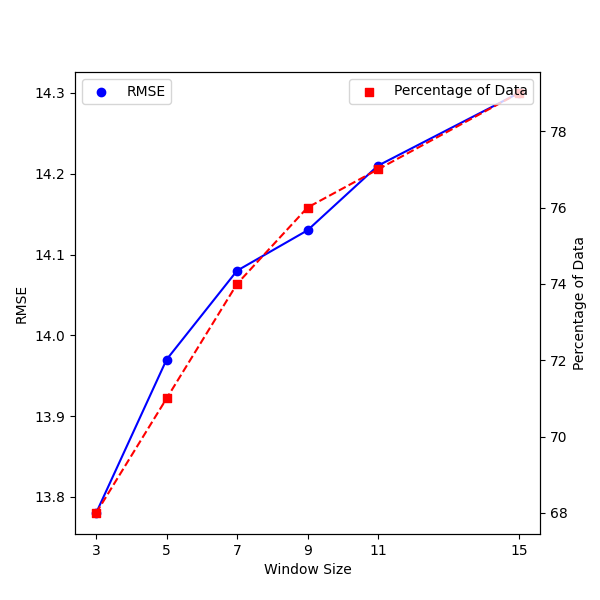}
		\caption{The influence of changing window size on the RMSE of regression as well as the percentage of missing AOD data}
		\label{ws}
	\end{center}
\end{figure}

As explained in Section \ref{aod_extraction}, first, AOD is extracted from the MAIAC file at each monitoring station. For this aim, different windows sizes, 3$ \times $3; 5$ \times $5; 7$ \times $7; 9$ \times $9; 11$ \times $11; 15$ \times $15, were experimented, and the effect of window size on estimating PM2.5 was evaluated. A univariate linear regression model was applied to evaluate the correlation between the extracted MAIAC AODs and corresponding PM2.5 values. Based on the results illustrated in Tab. \ref{Tab.ws}, increasing the size of the window degrades the accuracy of the univariate regression model. The best results were achieved using a 3$ \times $3 window size. However, the smaller window size boosts the chance of encountering missing values or poor quality AODs according to the criteria explained in Sections \ref{aod_extraction} and \ref{quality_extraction}. Fig. \ref{ws} shows that increasing the window size reduces the regression performance (raising RMSE values), whereas the percentage of possibly available AODs (non-missing values) is raised. From the slope of the RMSE plot, one can conclude that the degradation of performance dramatically changes by increasing the window size from 3$ \times $3 to 9$ \times $9 and larger sizes. Also, the percentage of data, shown by the red line-square plot, has the greatest change by varying the window size from 3$ \times $3 (nearly 67\%) to 7$ \times $7 (almost 74\%).  Nevertheless, Increasing the window size for AOD extraction at the monitoring stations causes mixing of the AOD values that belong to nearby air quality monitoring stations, which are located at a distance less than half of the window size. Another important aspect of exploring the optimal window size is the computational cost of AOD extraction. Increasing the window size requires more computational loads which can be problematic in the big data processing. 
In conclusion, 3$ \times $3 window size is selected as optimal window size for AOD extraction from MAIAC products in the study area.

\begin{table}[bt!]
	\centering \footnotesize
	\caption{The impact of changing window sizes on the correlation between PM2.5 values and MAIAC AODs}
	\label{Tab.ws}
	\begin{tabular}{l ccc}
		Window Size & \thead{RMSE \\  $\frac{ \mu  g}{ m^{3}}$}& \thead{MAE \\  $\frac{ \mu  g}{ m^{3}}$} & $ R^{2} $   \\\hline
		\toprule
		
		3$ \times $3 & 13.78&	10.91&	0.40  \\ 
		5$ \times $5 & 13.97&	11.01&	0.40 \\
		7$ \times $7 & 14.08&	11.08&	0.40 \\
		9$ \times $9 & 14.13&	11.11&	0.40 \\
		11$ \times $11 & 14.21&	11.17&	0.40 \\
		15$ \times $15 & 14.30&	11.21&	0.40 \\
		
		\hline		
	\end{tabular}	
\end{table}

\subsubsection{Influence of Quality of AODs on PM2.5 Estimation}\label{res.quality}

\begin{figure}[bt!]
	\begin{center}
		\includegraphics[width=1\columnwidth]{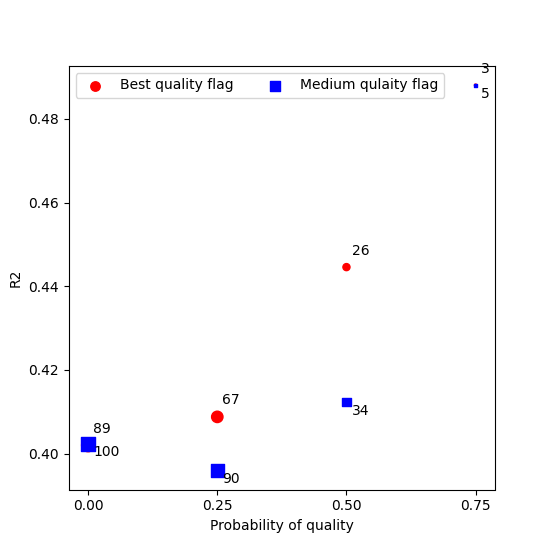}
		\caption{The influence of the probability of quality of AODs achieved based on the conditions 1 (medium) and condition 2 (best) on PM2.5 estimation}
		\label{fig.quality}
	\end{center}
\end{figure}

Fig. \ref{fig.quality}. displays the influence of quality of AOD values on predicting PM2.5. It should be noted that the simple linear regression model was also used for discovering the effect of AOD quality on the AOD-PM2.5 relationship.  The probability of quality for each extracted AOD was computed according to conditions 1 and 2 described in Section \ref{quality_extraction}. The zero probability means no quality condition was assumed for AODs inputting into the regression model and probability of 0.75 implies that at least 0.75\% of AODs within the extracting window comply with either condition 1 or 2. As shown in the figure, choosing AODs with the highest probabilities, i.e. highly qualified AOD values, can accurately estimate PM2.5 values. However, using conditions, in particular, condition 2, causes missing those AODs that could be beneficial for AOD-PM2.5 modeling, especially when applying sophisticated machine learning algorithms. Thus, instead of filtering based on the conditions, which are mostly helpful for simpler models like the univariate model, this investigation suggests using the probabilities exploited from AOD data as input features in machine learning-based modeling. In Section \ref{res.feature_importance}, it will be revealed that the probabilities can be imported as informative features along with meteorological data for PM2.5 estimation. 

\subsubsection{PM2.5 Outlier Removal}\label{res.outlier}
As explained in Section \ref{sec.outlier}, two strategies were applied for the detection and removal of PM2.5 outliers. Tab. \ref{Tab.outlier} represents results of univariate linear regression on data that have been modified using the aforementioned outlier removal strategies. The results illustrate that using the IQR technique can outperform the 3$ \sigma $ strategy in detecting outliers. Thus, the IQR strategy is chosen for outlier removal and data cleaning.

\begin{table}[bt!]
	\centering \footnotesize
	\caption{The effect of using IQR and 3$ \sigma $ strategies for outlier detection and removal from PM2.5 values. The univariate regression was performed to evaluate each outlier removal strategy. }
	\label{Tab.outlier}
	\begin{tabular}  {l ccc}
		Method & \thead{RMSE \\  $\frac{ \mu  g}{ m^{3}}$}& \thead{MAE \\  $\frac{ \mu  g}{ m^{3}}$} & $ R^{2} $   \\\hline
		\toprule
		IQR & 13.78	& 10.91	& 0.40 \\ 
		3$ \sigma $ & 14.70	& 11.49	& 0.43 \\

		\hline		
	\end{tabular}	
\end{table}

\subsubsection{Impact of Meteorological Data on Estimating PM2.5}\label{res.met}

Adding meteorological observations as input features is beneficial for PM2.5 estimation. For the meteorological data used in this study, Tab. \ref{Tab.krig} illustrates the best parameters tuned for kriging interpolation of each meteorological parameter. In other words, the mentioned settings give the best results for each category of meteorological data. After preparing meteorological data, they will be used along with AOD and other features (listed in Tab. \ref{feature}) for PM2.5 modeling.

\begin{table}[bt!]
	\centering \footnotesize
	\caption{The hyperparameters achieved from grid search with cross-validation for kriging interpolation of meteorological data in this study}
	\label{Tab.krig}
	\begin{tabular}  {l cc}
		Meteorological Data &Type of Kriging & Semivariogram  \\\hline
		\toprule
		
		d2m&	universal&	spherical \\
		t2m&	universal&	spherical \\
		blh&	ordinary&	spherical \\
		lai\_hv&	ordinary&	spherical \\
		lai\_lv&	ordinary&	spherical \\ 
		sp&	universal&	power \\
		ws10&	ordinary&	spherical \\ 
		wd10&	ordinary&	spherical \\ 
		uvb&	ordinary&	spherical \\ 
		cdir&	ordinary&	spherical \\ 
		RH&	universal&	spherical \\

		\hline		
	\end{tabular}
	
\end{table}
The importance of meteorological data in PM2.5 estimation on predicting PM2.5 is presented in Tab. \ref{Tab.aodnormal}.  As illustrated in Tab. \ref{Tab.aodnormal}, using meteorological features can improve the correlation coefficient of PM2.5 estimation up to 0.61, while without using this information, the correlation coefficient is around 0.40. It should be noted that the correlation of 0.40 is also achieved by the univariate model using normalized AOD by PBLH, which is also a meteorological parameter obtained from ECMWF models. Other metrics such as RMSE and MAE also confirm the efficiency of adding the aforementioned meteorological data. In more detail, the importance of meteorological variables on estimating PM2.5 concentration will be discussed in Section \ref{res.feature_importance}.

\subsubsection{Results of Feature Selection} \label{res.feature_importance}
Fig. \textcolor{blue}{S1} displays the importance of applied features in this study for PM2.5 modeling using XGBoost. As shown in this figure, the highest priority is for planetary boundary layer height (“blh”). Next, the normalized AOD (“nAODm”) works as the most informative attribute for PM2.5 regression. The plot also demonstrates that relative humidity has an important impact on estimating PM2.5. The lowest significance is related to “lai\_lv”, “month”, and “Prob\_medm” and “cdir”, respectively.


To support the results of feature importance determination using XGBoost, the heatmap plot of the correlation matrix (based on absolute correlation values) of the input features and the target variable (“PM$ _{c} $”) is displayed in Fig. \textcolor{blue}{S2}. According to the heatmap plot, “PM$ _{c}” $ has the highest correlation with “nAODm”, “RH”, “blh”, “lai\_hv”, “t2m”, “wd10”, “uvb”, and “ws10”, which has been also recognized as very important features by XGBoost. Some features such as “cdir” are significantly correlated with “wd10”. Thus, “wd10” can be a substitute for “cdir” in practice. Some features such as positional features (lat, long) have been identified as highly significant features by XGBoost, whereas, they have been recognized as less important attributes by correlation matrix. The main reason is that the correlation matrix is formed based on the linear correlation of attributes, while it is possible that “PM$ _{c} $” may not necessarily have a linear correlation with some features such as positional attributes. 


For more experiments, Fig. \textcolor{blue}{S3} displays the results of XGBoost performance (RMSE) with different settings associated with the presence of features as input variables in the process of regression. The applied settings for features have been presented in Tab \ref{Tab.setting}. The first setting is the removal of the least important feature i.e., “lai\_lv”. The plot shows that the performance of algorithms is slightly promoted. The removal of less important features is continued according to different settings presented in Tab. \ref{Tab.setting}. The red dashed line is the performance of the algorithm when employing all defined input features. As the blue plot depicts, the performance of the algorithm when using all variables is as same as the case of removing “lai\_lv”, month, and “Preb\_med” features.  This means that the mentioned features are useless in the process of XGBoost regression. Even, removal of “lai\_lv”, and month (setting S2) can slightly improve the algorithm performance. However, according to the plot of RMSE in Fig. \textcolor{blue}{S3}, removing more features based on settings such as S4, S5, and others degrades the algorithm accuracy. 


\begin{table}[t!]
	\centering \footnotesize
	\caption{Different settings of embedding and discarding of input features in the process of XGBoost regression}
	\label{Tab.setting}
	\begin{tabularx}{\columnwidth}  {l X}
		Settings & Discarded features   \\\hline
		\toprule
		S1&	lai\_lv	 \\ 
		S2&	lai\_lv + month	 \\
		S3&	lai\_lv + month + Preb\_med	 \\
		S4&	lai\_lv + month + Preb\_med + cdir	 \\
		S5&	lai\_lv + month + Preb\_med + cdir + sp	 \\
		S6&	lai\_lv + month + Preb\_med + cdir + sp + Prob\_best	 \\
		S7&	lai\_lv + month + Preb\_med + cdir + sp + Prob\_best + ws10	 \\
		S8&	lai\_lv + month + Preb\_med + cdir + sp + Prob\_best + ws10 + wd10 \\
		\hline		
	\end{tabularx}	
\end{table}

\subsection{Regression Modeling Results}\label{sec.reg_result}
After feature selection, different machine learning techniques were applied to model the relationship between PM2.5 and AOD, and other input features. Each algorithm has unknown parameters or hyperparameters that should be tuned to reach the best performance. The proper values of hyperparameters were determined during the training (70\% of the entire data). After that, the independent data (30\% of the whole information) as unseen data (also called test data) were employed to evaluate the efficiency of algorithm. In this study, the 5-fold cross-validation strategy was used for training all algorithms except deep learning methods. Since training of DAE and DBN demands a lot of training data and also spends a large deal of computing time, the initial training data was split into two sets as training (80\%) and remaining training data (20\% ) as validation. 

\begin{figure*}[bt!]
	\begin{center}
		\includegraphics[width=1\textwidth]{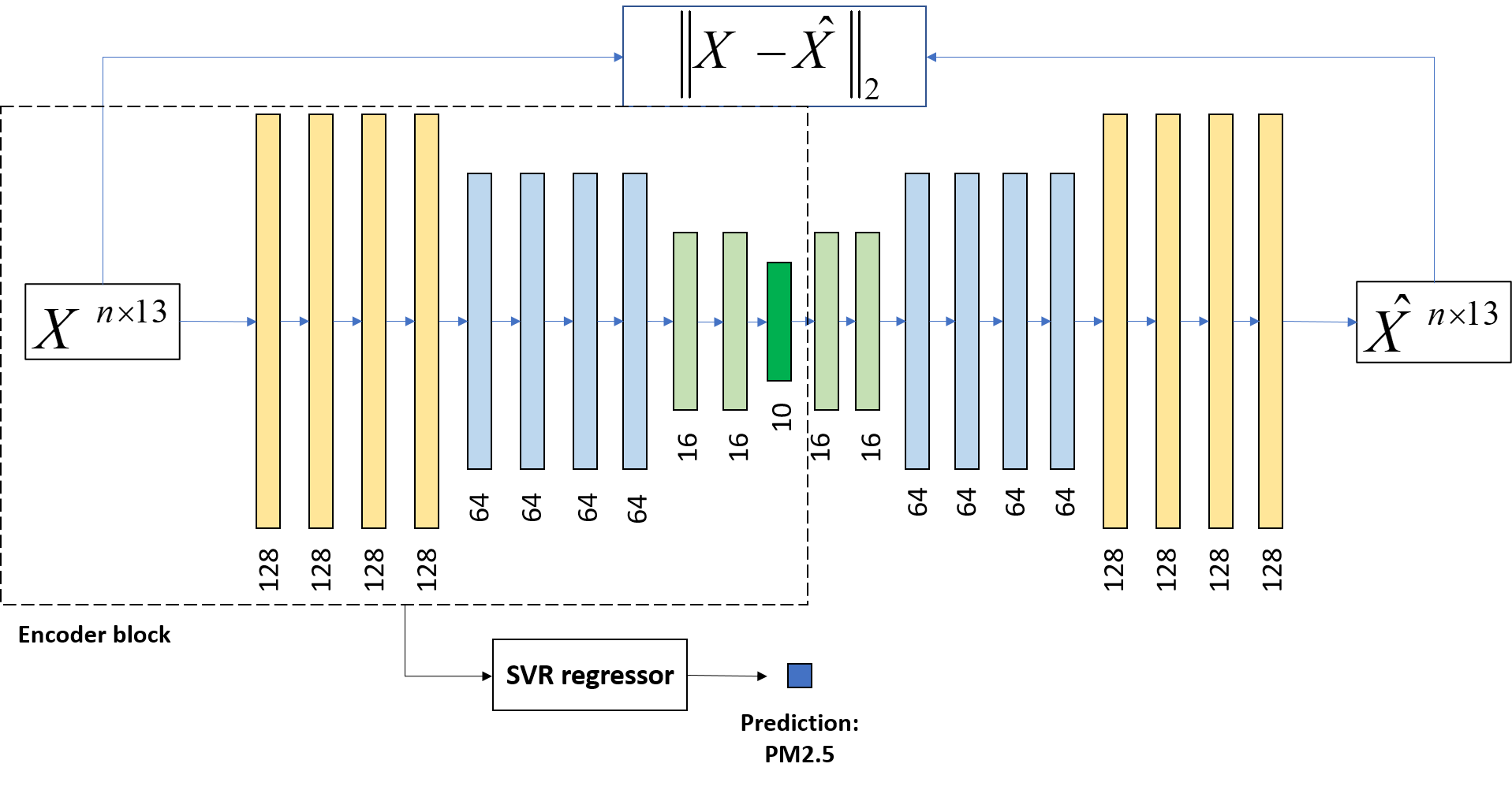}
		\caption{The structure of DAE used in this study for PM2.5 modeling}
		\label{AE}
	\end{center}
\end{figure*}

\begin{figure*}[bt!]
	\begin{center}
		\includegraphics[width=0.6\textwidth]{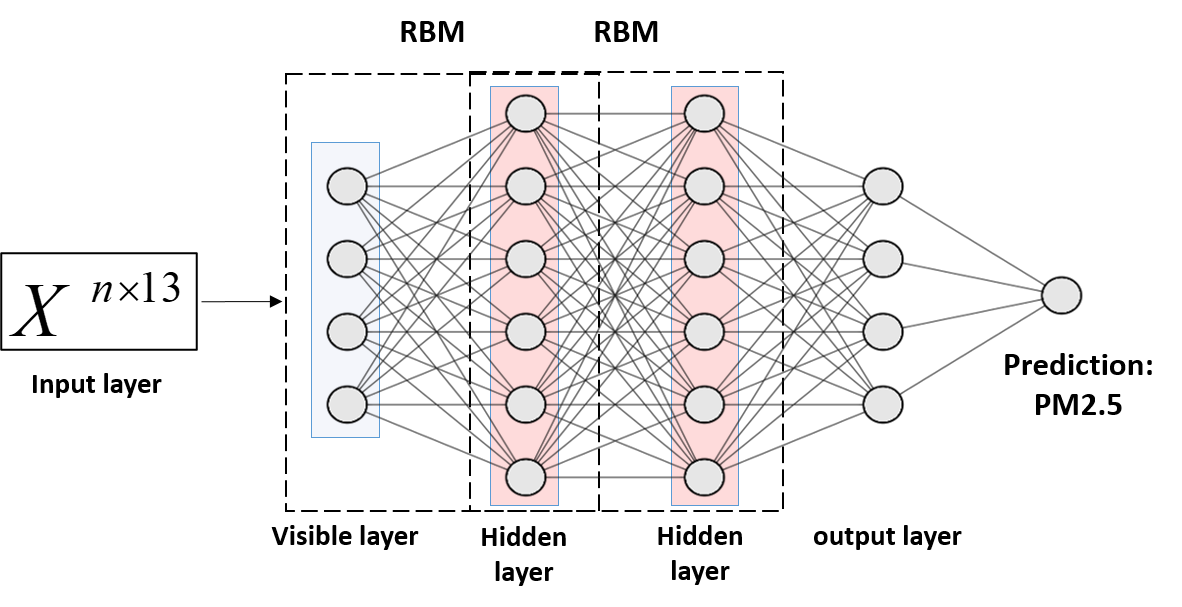}
		\caption{The structure of deep belief network (DBN) constructed by a stack of restricted Boltzmann machines (RBM) used for PM2.5 regression}
		\label{DBN}
	\end{center}
\end{figure*}

The structures of the designed deep autoencoder (DAE+SVR) and deep belief networks (DBN) employed for PM2.5 concentration using MAIAC AOD values and other input features in this study have been displayed in Fig. \ref{AE} and \ref{DBN}, respectively.

Other hyperparameters such as learning rate, optimization method, etc., were identified during the training using the validation data.
The hyperparameters tuned for machine learning algorithms used in this study have been represented in Tab. \ref{Tab.hyp}. After fine-tuning, the developed models were evaluated on train data and test data to give models' performances during training and testing, respectively.
Tab. \ref{Tab.model} collects performances of the applied machine learning techniques.  

\begin{table}[bt!]
	\centering \footnotesize
	\caption{Hyperparameter setting of machine learning algorithms used in this study}
	\label{Tab.hyp}
	\begin{tabularx}{1\columnwidth}{l X}
		Algorithm& Main hyperparameters \\\hline
		\toprule
		Univariate&	 ---\\
		Multivariate&	 ---\\
		Ridge&	 regularization parameter: 0.1\\
		Lasso&	 regularization parameter: 0.1\\
		SVR&  kernel type: RBF, regularization parameter: 100, epsilon: 0.1\\
		Random Forest&	 No. of estimators: 500, max depth: 10, max features: 0.5, min samples in a leaf: 1,  criterion: MSE\\
		Extra Trees & No. of estimators: 500, max depth: 10, max features: 0.8, min samples in a leaf: 1, criterion: MSE\\
		XGBoost& booster: decision tree, No. of trees: 2000, criterion: MSE, learning rate: 0.3, maximum depth: 6, max features: 1, min child weight: 1, Gamma: 0, \\
		DBN & No. of hidden layers: 2 (64, 10 neurons), learning rate of RBM: 0.01, learning rate of the network: 0.001, optimizer: SGD, No. of epochs for training RBM: 50, No. of backpropagation iteration: 200, mini-batch: 256, activation function: ReLU, loss: mse\\
		DAE& structure: Fig. \ref{AE}, optimization: Adam, learning rate: 0.001, activation: ReLU, loss: mse, regularization: $l_{2}$ weight penalty with factor of 0.001, epoch: 200) + SVR (kernel: RBF, regularization: 100, epsilon: 0.1)\\
	\end{tabularx} 	
\end{table}

\begin{table*}[tb]
	\centering \footnotesize
	\caption{The performances of different machine learning techniques used in this study for PM2.5 estimation.}
	\label{Tab.model}
	\begin{tabularx}{\textwidth}{XX ccc ccc}
		
		&  & \multicolumn{3}{c}{Model training}& \multicolumn{3}{c}{Model testing} \\
		Category&Method & RMSE  $\frac{ \mu  g}{ m^{3}}$& MAE  $\frac{ \mu  g}{ m^{3}}$ & $ R^{2} $ & RMSE  $\frac{ \mu  g}{ m^{3}}$ & MAE $\frac{ \mu  g}{ m^{3}}$ & $ R^{2} $ \\
		\toprule
		\multirow{4}{*}{Linear Methods} & Univariate &	13.82&	10.93&	0.40&	15.45&	12.06&	0.41 \\
		& Multivariate &	10.96&	8.64&	0.61&	12.38&	9.84&	0.59 \\
		& Ridge &	10.92&	8.60&	0.61&	12.26&	9.74&	0.59 \\
		& Lasso &	11.18&	8.78&	0.59&	11.74&	9.29&	0.58 \\
		\hline
		\multirow{1}{*}{Kernel Methods} &SVR & 10.27&	7.88&	0.63&	10.36&	7.98&	0.63 \\
		\hline
		\multirow{3}{*}{Ensemble Methods}& Random Forest&	7.85&	6.13&	0.85&	9.51&	7.50&	0.69 \\
		&Extra Trees&	8.61&	6.80&	0.80&	9.63&	7.66&	0.68 \\
		&XGBoost&	\textbf{6.39}&	\textbf{4.79}&	\textbf{0.92}&	\textbf{8.97}&	\textbf{6.88}&	\textbf{0.74} \\
		\hline
		\multirow{2}{*}{Deep learning} &DBN &	9.61&	7.46&	0.70&	9.99&	7.67&	0.66 \\
		&DAE + SVR&	7.97 &	6.03&	0.83&	9.75&	7.32&	0.68 \\

	\end{tabularx} 	
\end{table*}

According to Tab. \ref{Tab.model}, the best results are achieved using XGBoost according to the model performance on test data. The RMSE and MAE of XGBoost are 6.39 and 4.79 on train data and 8.97 and 6.88 on test data, respectively. Random Forest was trained well on train data (RMSE = 7.85, MAE= 6.13, and $R^{2} $= 0.85), However, its accuracy decreased when it was applied to test data. After that,  
DAE+SVR with RSME of 9.75, MAE of 7.32, and $R^{2} $ of 0.68 gives the best results on test data. Another deep neural network structure i.e., DBN could also outperform SVR and linear methods, while its accuracy is less than extra trees. Among linear methods as most straightforward regression techniques, linear ridge regressor has the highest accuracy, which shows the efficiency of regularization. 
Among the models, the lowest accuracy is related to the univariate models. 
Since the univariate regression model does not employ other valuable features, it leads to less performance than other models. 

\subsection{Deployment Results}
The achieved model from the regression stage was finally employed for estimating PM2.5 in locations that were not sensed by ground sensors. Considering the overall performance of developed models on both train and test data, the tuned XGBoost model was selected as the final model for PM2.5 map generation. 

In the process of regression modeling, the missing values are not involved since sufficient valid samples are available for any type of regression algorithms.  However, it is critical in the deployment stage, where the goal is to produce a raster of PM2.5 estimates. In the deployment stage, instead of interpolation of missing AOD values, the achieved model from the output of the regression phase is applied to estimate PM2.5 using valid AODs. Then, the locations with invalid  AODs are directly filled by PM2.5 obtained through interpolation of PM2.5 estimates output of the trained regression model.

\begin{figure*}[t!]
	\centering
	\subfloat[locations of ground stations as well estimated PM2.5 by the developed model]{%
		\includegraphics[width=0.49\linewidth]{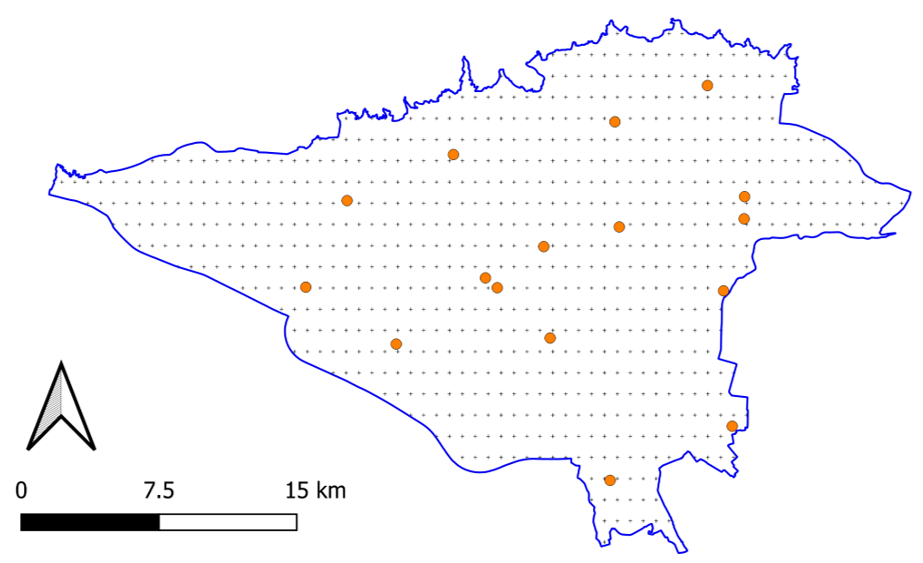}}
	\label{fig.2018001point}
	\subfloat[Jan. 1, 2018: “Unhealthy”]{%
		\includegraphics[width=0.49\linewidth]{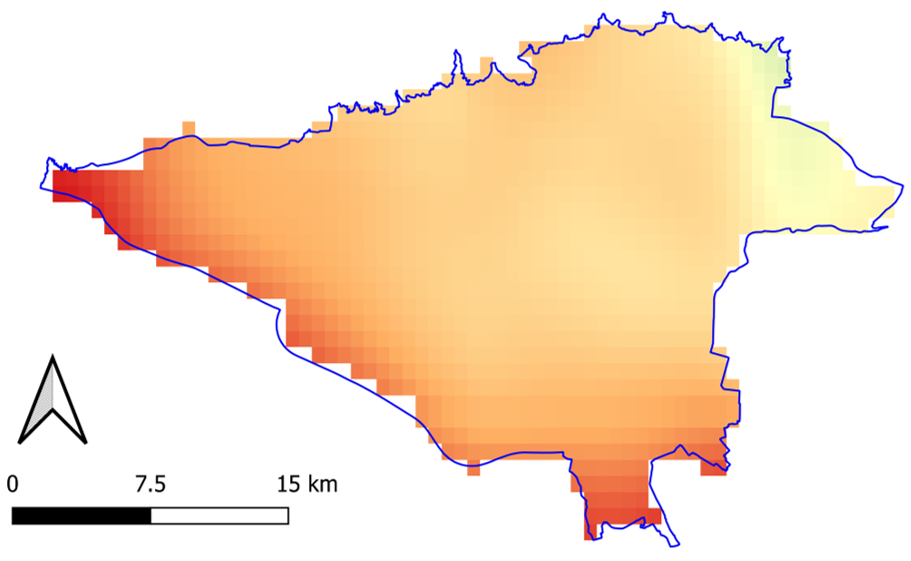}}
	\label{fig.2018001}
	\subfloat[Jan. 2, 2018: “Unhealthy for sensitive group”]{%
		\includegraphics[width=0.49\linewidth]{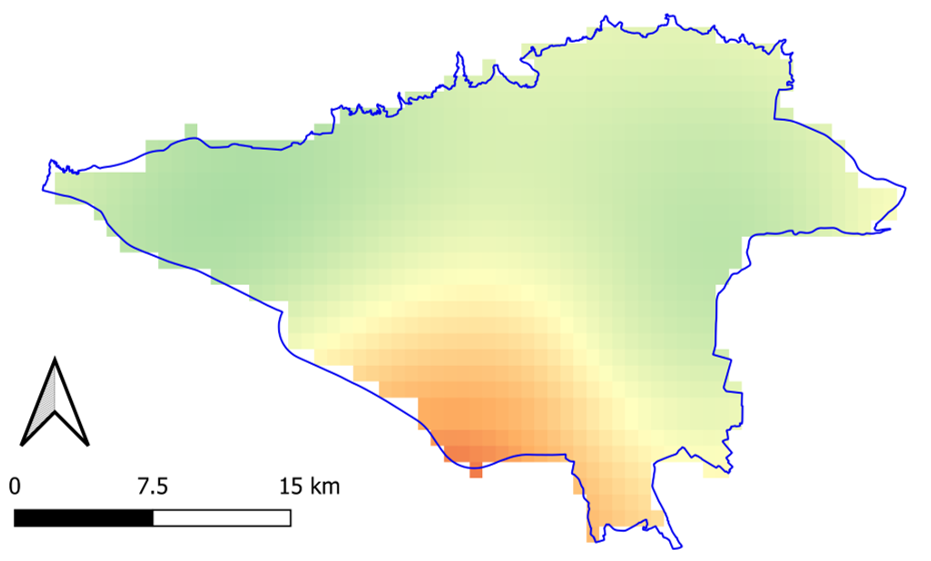}}
	\label{fig.2018002}
	\subfloat[Jan. 3, 2018: “Moderate”]{%
		\includegraphics[width=0.49\linewidth]{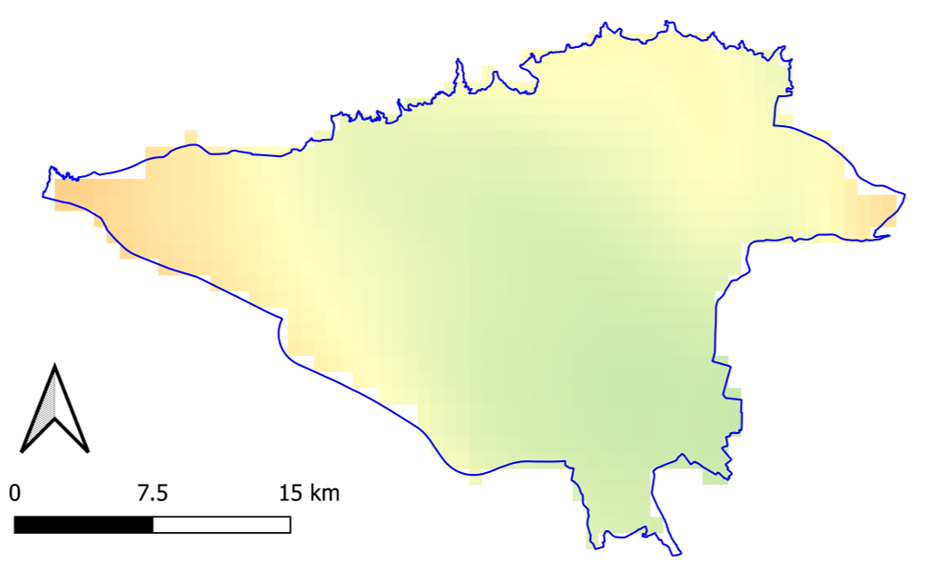}}
	\label{fig.2018003}
	\subfloat[Feb. 25, 2018: “Clean”]{%
		\includegraphics[width=0.49\linewidth]{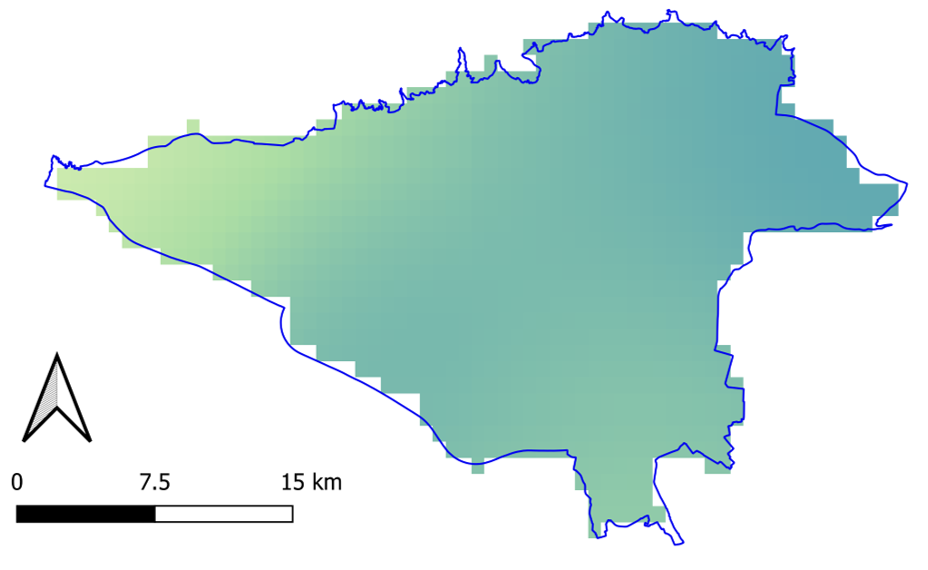}}
	\label{fig.2018056}
	\subfloat[Legend]{%
		\includegraphics[width=0.2\linewidth]{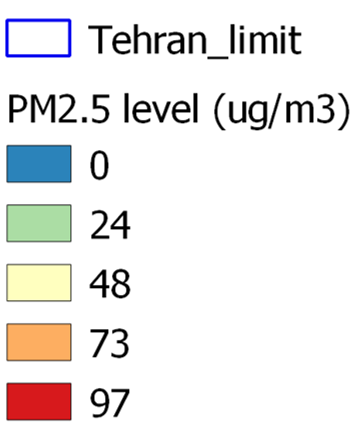}}
	\label{fig.leg}
	
	\caption{a) locations of predicted PM2.5 (black crosses) and ground stations (orange circles), b-d) Visualization of high resolution Pm2.5 map generated by the developed regression model (XGBoost), 
	}
	\label{fig.PMmap} 
\end{figure*}

Fig. \ref{fig.PMmap}.a depicts ground stations (orange circles) as well as locations of those PM2.5 values have been generated by the developed XGBoost model (black crosses). The produced PM2.5 data can be supposed as new PM2.5 measuring stations (quasi-stations) and thus be combined with ground stations to indicate PM2.5 variations for higher resolution mapping better.
 Finally, a high resolution daily map of PM2.5 is produced using an interpolation technique such as kriging using all estimated and observed PM2.5 values.
 
 Fig. \ref{fig.PMmap} displays exemplary high resolution (1 km) maps produced by the developed machine learning model on four different dates. Four dates were selected based on the versatile levels of pollution over the city reported by  Tehran's Air Quality Control Company (AQCC). The dates are Jan. 1, 2018, that was announced “Unhealthy” based on the air quality index (AQI), Jan. 2, 2018, as “Unhealthy for sensitive group”, Jan. 3, 2018, as “Moderate”, and Feb. 25, 2018, as a “Clean” day. The efficiency of the proposed framework for each pollution level based on the reported AQI is shown by generated PM2.5 maps.

As shown in Figs. \ref{fig.PMmap}, the PM2.5 map of Jan. 1th illustrates that the most areas have PM2.5 levels of more than 73 $\frac{ \mu  g}{ m^{3}}$ that also confirms the level of pollution “Unhealthy” on that date. The second of Jan. is “Unhealthy for sensitive people”, which can also be inferred from the produced map. On this date, the southwest of the city is still suffering from the high amount of PM2.5; however, the concentration of PM2.5 in most areas is less than the previous day. On the third day, based on the achieved map, the level of PM2.5 in a vast part of the city (except west and northeast) is almost less than 24, which means decreasing the level of pollution to moderate as reported by AQCC. Finally, Fig.\ref{fig.PMmap}.e displays that the most estimated PM2.5 data have small values representing a “Clean” day.

From the maps presented in Fig. \ref{fig.PMmap} as well as results provided in Tab. \ref{Tab.model}, it can be concluded that the developed framework including different stages can be successfully employed for daily high resolution map generation of PM2.5 over Tehran.

 Finally, the results of this investigation were compared to previous studies implemented in Tehran for PM2.5 estimation. As compared in Tab. \ref{Tab.literature}, while this investigation could successfully lead to daily, 1 km mapping of PM2.5 in Tehran, previous studies, in the best situation, could only produce a 3 km resolution map with accuracy less than that was achieved in this paper. 
\begin{table*}[bt!]
	\centering \footnotesize
	\caption{Comparing results of this study with previous implementations reported in the literature}
	\label{Tab.literature}
	\begin{tabular}{l ccccccc}
		Study& Time &Data, Resolution & Model&RMSE  $\frac{ \mu  g}{ m^{3}}$& MAE  $\frac{ \mu  g}{ m^{3}}$ & $ R^{2} $ & Daily PM2.5 MAP \\\hline
		\toprule
		This study & 2013-2019& MAIAC-MODIS, 1 km & XGBoost & 8.97&	6.86&	0.74& 1 km PM2.5 map\\
		\citet{atmos10070373}&2015-2018 & DB-DT-MODIS, 3 km& XGBoost& 15.15 & 10.94& 0.67 & Not reported\\
		\citet{NABAVI2019889}& 2011-2016& MAIAC-MODIS, 1 km& Random Forest& ---& ---& $ < $ 0.50 & Seasonal map (1 km)\\
		\citet{GHOTBI2016333}& March to Nov. 2009& DT-MODIS, 3 km & WRF-Multivariate & 16.91 & ---& 0.73 & 3 km PM2.5 map\\

	\end{tabular} 	
\end{table*}

\section {Conclusion}\label{sec.conclusion}
This paper investigated the possibility of PM2.5 estimation using the MAIAC AOD data and meteorological information over Tehran. For this aim, a framework including three main stages, data preprocessing; regression modeling; and model deployment for generating high resolution map of PM2.5 was proposed. During the data preprocessing, the effect of several factors and parameters on PM2.5 estimation such as window size for AOD extraction, the impact of AOD normalization, the significance of adding meteorological data, the role of involving AOD quality was evaluated. Regression modeling was performed from different categories of machine learning techniques for estimating PM2.5 using input features. Model performance results illustrated that the decision tree ensemble approaches such as random forest and XGBoost were the best choice for PM2.5 estimation from AOD and meteorological data. The developed regression model was finally employed for producing the 1 km resolution PM2.5 concentration maps, which could be potentially exploited for monitoring, predicting the air quality condition, and also detecting main air pollution sources. Inspection of generated maps on exemplary days with different levels of pollution based on the officially reported air quality index by AQCC of Tehran confirmed the efficiency of the developed framework.

In the future, more attempts will be conducted to handling the challenges in the process of high resolution map generation such as involving other effective features in PM2.5 modeling, imputation of missing AOD values and improving the performance of modeling by developing more advanced machine learning techniques.

\section{Acknowledgments}
The author wants to thank everyone who has provided the required data
for this research, Tehran's Air Quality Control Company (AQCC) for the ground PM2.5 measurements; NASA EarthData for the MAIAC MODIS products; and ECMWF for the meteorological data. 

\section{Appendix A. Supplementary data}
Supplementary data to this article can be found online at \textcolor{red}{xxx}

\bibliographystyle{model5-names}
\biboptions{authoryear}
\bibliography{sample.bib}

\end{document}